\documentclass[10pt,conference,a4paper]{IEEEtran}
\usepackage{amsmath}
\usepackage{times}
\usepackage{graphicx}
\usepackage{color}
\usepackage[noadjust]{cite}  
\usepackage{listings}
\usepackage{amssymb}
\usepackage{supertabular}
\usepackage{longtable}
\usepackage{enumerate}
\usepackage[table]{xcolor}
\usepackage{array} 
\usepackage{float}
\usepackage{caption}
\usepackage{mathtools}
\usepackage{supertabular}
\usepackage{longtable}
\usepackage{url}
\usepackage{multirow}
\usepackage{booktabs}

\usepackage{color}
\usepackage{comment}
\usepackage{mathrsfs}  
\usepackage{amsfonts, amssymb}
\usepackage{float}
\usepackage[font=small,skip=2.5pt]{caption}
\ifCLASSOPTIONcompsoc
  \usepackage[nocompress]{cite}
\else
\fi

\ifCLASSINFOpdf
\else
\fi

\begin{document}
\title{Regularized Flexible Activation Function Combinations for Deep Neural Networks}

\author{\IEEEauthorblockN{Renlong Jie, Junbin Gao, Andrey Vasnev, Min-ngoc Tran}
\IEEEauthorblockA{Discipline of Business Analysis, School of Business\\
The University of Sydney\\
Camperdown NSW 2006\\
Email: renlong.jie@sydney.edu.au, junbin.gao@sydney.edu.au, 
andrey.vasnev@sydney.edu.au, minh-ngoc.tran@sydney.edu.au
}
}

\maketitle

\begin{abstract}
Activation in deep neural networks is fundamental to achieving non-linear mappings. Traditional studies mainly focus on finding fixed activations for a particular set of learning tasks or model architectures. The research on flexible activation is quite limited in both designing philosophy and application scenarios. In this study, three principles of choosing flexible activation components are proposed and a general combined form of flexible activation functions is implemented. Based on this, a novel family of flexible activation functions that can replace sigmoid or tanh in LSTM cells are implemented, as well as a new family by combining ReLU and ELUs. Also, two new regularisation terms based on assumptions as prior knowledge are introduced. It has been shown that LSTM models with proposed flexible activations P-Sig-Ramp  provide significant improvements in time series forecasting, while the proposed P-E2-ReLU achieves better and more stable performance on lossy image compression tasks with convolutional auto-encoders. In addition, the proposed regularization terms improve the convergence, performance and stability of the models with flexible activation functions. The code for this paper is available at \url{https://github.com/9NXJRDDRQK/Flexible_Activation}.
\end{abstract}

\IEEEpeerreviewmaketitle

\section{Introduction}\label{Sec:1}
Deep learning is one of the most powerful techniques in modern artificial intelligence \cite{lecun2015deep}. One reason is its ability in approximating complex functions with a large but limited number of parameters \cite{cybenko1989approximation, hornik1991approximation}, while the regular layer structures make it possible to be trained with efficient back propagation algorithms \cite{goodfellow2016deep}. 

In a deep neural network, the weights and bias take account of linear transformation of the data flow, while the activation functions bring in non-linearity. It is remarked in \cite{hornik1991approximation} that activation functions do not perform equally well if we take minimal redundancy or computational efficiency into account. Thus the selection of activation function for different tasks is a critical issue in a sense. Traditionally, people train the weights of linear transformations between layers while keeping the activation functions fixed, and usually one identical activation function is used for all the neurons on each single layer. For example, the rectifier linear units (ReLU) are used as the default choice for the activation in hidden units for feed forward neural networks and in a large proportion of convolutional neural networks \cite{nair2010rectified}, while sigmoid and tanh functions are used where output values are bounded, such as in output layers for classification problems and the gate activations in recurrent cells \cite{gers1999learning, chung2014empirical}. 

The drawback of ReLU activation is the issue of dead unit when the input is negative, which motivated the introduction of functions with non-zero values in the negative range, including leaky-ReLU and exponential linear units (ELU) \cite{maas2013rectifier,clevert2015fast}. By considering other properties that could be helpful in training, such as dropout or self-regularization effects, more types of activation functions such as GeLu and Mish are proposed \cite{hendrycks2016gaussian, misra2019mish}. There are also activation functions designed for particular learning tasks such as reinforcement learning \cite{elfwing2018sigmoid}, where the activation functions used for neural network function approximation are computed by the sigmoid function multiplied by its input. Moreover, reinforcement learning can be implemented for searching activation functions with an RNN controller \cite{ramachandran2017searching}. Beyond properly choosing the activation functions, techniques such as clipping and batch normalization can be implemented to make activations robust to explosion or vanishing gradients\cite{ioffe2015batch,lin2017deep}. Traditionally, explosion or vanishing gradients in back propagation are also issues that harm the performance of model largely due to the shape of activations \cite{bengio1994learning,hochreiter1998vanishing,pascanu2012understanding}. Beyond properly choosing  the activation functions, techniques such as clipping and batch normalization can be implemented to alleviate these issues to some extent \cite{ioffe2015batch,lin2017deep}. 

With a large enough neural network and sufficient training time, the model can effectively learn the patterns from data with possible high accuracy, however it is not straightforward to confirm that learning process is the most efficient and the results are the most accurate. One possible solution for accelerating model training is to introduce flexible or trainable activation functions \cite{agostinelli2014learning,he2015delving,chung2016deep}. Even though this requires higher computing and storing cost that is proportional to the number of neurons, the performance of models can be improved in a more efficient manner than increasing the number of basic model parameters or the number of neurons.
For example, as the leaky-ReLU function has a hyper-parameter to be optimized, which is the slope of its negative part, parameterized Relu (PReLU) was proposed to make this slope adapt to the data within specific neurons and to be learned during the training process \cite{he2015delving}. Meanwhile, \cite{li2018improving} proposes the parameterized version of ELU activation, which introduces two parameters to control the shape of exponential curve in the negative region.    

It can also be a blending of different commonly used activations, where the trainable parameters are the weights for the combination components \cite{sutfeld2018adaptive, manessi2018learning}. Since different activation functions can have very similar behavior in some specific regions, a more generative way is to consider their Taylor expansions at $0$ point and use a weighted combination of polynomial functions with different orders instead \cite{chung2016deep}. For containing those functions that are not centered at $0$, one choice is to train a piece-wise function adaptively \cite{agostinelli2014learning}. The similar effect can be achieved by Maxout activation, which is quite helpful in promoting the efficiency of models with dropout \cite{goodfellow2013maxout}. Beyond that, there are also studies on making the most of the non-linear properties by introducing adaptation mechanism on the Softmax layers \cite{flennerhag2018breaking}, which achieve the former state-of-the-art results on several natural language processing (NLP) tasks. 

The limitation of existing studies can be illustrated as follows. First, most of existing work focus on some specific forms of parameterized activation functions rather than a more general form, or consider each component of the combination as a fixed activation function. Second, there is a lack of study on flexible activations with bounded domain such as sigmoid and tanh. Third, existing works rarely discuss the regularization on activations parameters, which have different nature from normal model parameters.
In this study, we consider the activation function as a combination of a set of trainable functions following the constraints of several principles. Based on these principles, we develop two flexible activation functions that can be implemented for bounded or unbounded domain.
In addition, layer-wise regularization on activation parameters is introduced to reduce the variance caused by activation functions. 

Section ~\ref{Sec:2} is an introduction of the main idea and methodology of this paper with initial theoretical analysis. Section ~\ref{Sec:3} is the experimental results of our methods, in which we compare the performances of models with newly proposed flexible activation functions and baseline models on several datasets. Also we check the effect of regularization on activation parameters. Section ~\ref{Sec:4} is the discussion on number of parameters and computational complexity. Section ~\ref{Sec:5} is the conclusion.

\section{Methodology} \label{Sec:2}
Existing studies on trainable activation function mainly focus on the case where the output is unbounded. However, activation functions with bounded domain, such as sigmoid and tanh, are implemented in a large number of models. One scenario is recurrent neural networks, such as LSTM, whose cell has the functional mapping as follows:   
\begin{equation}
\begin{split}
&\boldsymbol{f}_t = \sigma(\boldsymbol{W}_{fx}\boldsymbol{x}_t + \boldsymbol{W}_{fh}\boldsymbol{h}_{t-1}+\boldsymbol{b}_f)\\   
&\boldsymbol{i}_t = \sigma(\boldsymbol{W}_{ix}\boldsymbol{x}_t + \boldsymbol{W}_{ih}\boldsymbol{h}_{t-1}+\boldsymbol{b}_i)\\
&\boldsymbol{o}_t = \sigma(\boldsymbol{W}_{ox}\boldsymbol{x}_t + \boldsymbol{W}_{oh}\boldsymbol{h}_{t-1}+\boldsymbol{b}_o)\\
&\boldsymbol{g}_t = \tanh(\boldsymbol{W}_{gx}\boldsymbol{x}_t + \boldsymbol{W}_{gh}\boldsymbol{h}_{t-1}+\boldsymbol{b}_g)\\
&\boldsymbol{c}_t = \boldsymbol{f}_t * \boldsymbol{c}_{t-1} + \boldsymbol{i}_t * \boldsymbol{g}_t\\
&\boldsymbol{h}_t = \boldsymbol{o}_t * \tanh(\boldsymbol{c}_t)
\end{split} \label{Eq:3}
\end{equation}
where the cell structure includes multiple sigmoid and tanh activation functions. If we want to introduce trainable ones in the same positions, the output domains of the newly introduced flexible activation functions should be the same with the original fixed ones.
\subsection{Parameterized Activation Function Combinations} \label{Sec:2.1}
In this study, we implement a general form of parameterized activation functions linearly combined by different activation functions as components\cite{sutfeld2018adaptive, manessi2018learning}. We further extend the existing works by introducing trainable parameters in each component, which is essential for building trainable activations with bounded domain. We assume that the parameters in the combined activation functions can be different for each neuron, which can be trained during the main training process of the model parameters with back propagation.
\begin{equation}
\begin{split}
    o_i(z, \boldsymbol{\alpha}^i, \boldsymbol{\beta}^i) &= \sum_{k=1}^K \alpha_{ik}
    f_k(z, \boldsymbol{\beta}_{ik}),\\
    \quad\sum^K_{k=1}\alpha_{i,k} &= 1,\quad 0\leq\alpha_{i,k}\leq 1\quad \forall k,i
    \end{split}\label{Eq:1}
\end{equation}
where $i$ indexes the neuron, and $z = z_l =\boldsymbol{W_l}X_{l-1}+\boldsymbol{b_l}$ is the input of the activation layer indexed by $l$. This means that, at each neuron $i$, it is possible to have its own set of parameters $\boldsymbol{\alpha}^i =[\alpha_{i1}, ..., \alpha_{iK}]^T$ and $\boldsymbol{\beta}^i = [\boldsymbol{\beta}_{i1},..., \boldsymbol{\beta}_{iK}]$ where $\alpha_{ik}$ is the combination weights and $\boldsymbol{\beta}_{ik}$ is the activation parameter vector for the $k$-th component activation $f_k$, respectively. Thus Eq.~\eqref{Eq:1} defines a form of activation function as a linear combination of a set of basic parameterized non-linear activation functions $f_k(z, \boldsymbol{\beta}_k)$ with the same input $x$ to the neuron. Normally, we require $0\leq\alpha_{i,k}\leq 1$ for all $k$ and $i$ to ensure that the output is strictly bounded between 0 and 1. This setting will take advantage of the low computational costs of existing activation functions, while it will be much easier to implement weights normalization when we need a bounded activation function. 

Since the specific activation function corresponding to each neuron only depends on its own activation parameters, the back propagation of these activation parameters by stochastic gradient descent can be done as follows: 
\begin{equation}
\begin{split}
&\alpha_{ik} \rightarrow \alpha_{ik} - \gamma \frac{\partial L}{\partial \alpha_{ik}} 
= \alpha_{ik}-\gamma \frac{\partial L}{\partial o_i}\cdot f_{ik}(z, \boldsymbol{\beta}_{ik})\\
&\boldsymbol{\beta}_{ik} \rightarrow \boldsymbol{\beta}_{ik} - \gamma \frac{\partial L}{\partial \boldsymbol{\beta}_{ik}} 
= \boldsymbol{\beta}_{ik} - \gamma \frac{\partial L}{\partial o_i}\cdot \alpha_{ik} \frac{\partial f_{ik}(z, \boldsymbol{\beta}_{ik})}{\partial \boldsymbol{\beta}_{ik}}
\end{split} \label{Eq:2}
\end{equation}
where $L$ is the loss function of all parameters of the model, $i$ is the index of the hidden neuron with output $o_i$ and  $k$ is the index of combined flexible activation functions. Here we use a simplified expression that does not include the indices of layer and training examples in each mini-batch. $\gamma$ is the learning rate of gradient descent for all the parameters in activation functions. With the gradients given by $\partial L/\partial \alpha_{ik}$, adaptive optimizers such as AdaGrad, Adam and RMSProp can also be applied. In general, gradient descent approach and its derivatives can push the activation parameters toward the direction that minimizes the empirical risk of the model on training data. 

To build effective combinations with the general form given by Eq.~\eqref{Eq:1}, we introduce the following three principles for selecting the components:
\begin{itemize}
    \item \textbf{Principle 1}: Each component should have the same domain as the baseline activation function.
    \item \textbf{Principle 2}: Each component should have an equal range as the baseline activation function.
    \item \textbf{Principle 3}: Each component activation functions should be expressively independent of other component functions with the following definition.
\end{itemize}
\textbf{Definition 1}: If a component activation function $f_k$ is \textbf{expressively independent} of a set of other component functions: $f_1,...,f_n$, there does not exist a set of combination coefficients $\alpha_1$,...,$\alpha_n$, inner activation parameters $\beta_1$,...$\beta_n$, parameters of the previous linear layers $\boldsymbol{W'}$, $\boldsymbol{b'}$ such that for any input $X$, activation parameters $\boldsymbol{\beta_k}$, and parameters of the previous linear layer $\boldsymbol{W_k}$, $\boldsymbol{b_k}$, the following equation holds:
\begin{equation}
\begin{split}
&f_k(z_k,\boldsymbol{\beta_k}) = f_k(\boldsymbol{W_k} X+\boldsymbol{b}_k,\boldsymbol{\beta_k})\\
&= \sum_{i=1}^n \alpha_i f_i(\boldsymbol{W'} X +\boldsymbol{b'}, \boldsymbol{\beta_i})
= \sum_{i=1}^n \alpha_i f_i(z', \boldsymbol{\beta_i})
\end{split}
\end{equation}
\textbf{Proposition 1}: For a single-layer network with $m$ neurons, if a component activation function $f_k$, which is not expressively independent of other components, is excluded, we need at most $2m$ neurons to express the same mapping exactly.\label{Theorem:1}\newline
\textbf{Proof}: For any original combined activation function: $F = \sum_{i=1}^n 
\alpha_i f_i$. Assume that $f_k$ is not expressively independent. If we exclude $f_k$ from the combination and get $F'=\sum_{i=1, i\neq k}^n \alpha_i f_i$, then for any input $X$, parameters of the previous linear layer: $W$, $b$, and activation parameters: $\{\beta\}_{i=1}^n$ of a specific neuron, we have:
\begin{equation}
    F(WX+b, \beta) = \sum_{i=1, i\neq k}^n {\alpha}_i f_i(WX +b, \beta_i) + f_k(WX +b, \beta_k)
\end{equation}
and there exist $W'$, $b'$ and $\{\beta\}_{i=1, i\neq k}^n$ such that:
\begin{equation}
\begin{split}
    F(WX+b, \beta) &= \sum_{i=1, i\neq k}^n {\alpha}_i f_i(WX +b, \beta_i)\\
    &+ \sum_{i=1, i\neq k}^n {\alpha}^{'}_i f_i(W' X +b', \beta_i^{'})
    \end{split}
\end{equation}
which can be expressed by two neurons with function $\{f\}_{i=1, i\neq k}^n$ with corresponding weights and bias in the previous linear layer. Therefore, for a single-layer network with $m$ neurons, we need at most $2m$ neurons without $f_k$ to express the any mappings by the original function.

The first two principles are aiming at keeping the same ranges and domains of the information flow with the mapping in each layer. The third principle is aiming at reducing the redundant parameters that do not contribute to the model expressiveness even with limited number of units. For example, $\sigma_1(z) = 1/(1+e^{-\beta z})$ is not expressively independent with $\sigma(z) = 1/(1+e^{-z})$ since when $\boldsymbol{W'}=\beta \boldsymbol{W}$, we have $\sigma(\boldsymbol{W'}X) = \sigma_1(\boldsymbol{W} X)$. Therefore, the combined activation $a(z,\beta) = \alpha_1 \sigma(z)+ (1-\alpha_1) \sigma(\beta z)$ will not be a good choice. Based on this, we can then design the combined trainable activation functions for both bounded or unbounded domains.

\subsection{P-Sig-Ramp : Sigmoid/Tanh Function substitute with bounded domain} \label{Sec:2.2}

\begin{figure}[th] 
\begin{center}
 \includegraphics[width=1.0\linewidth]{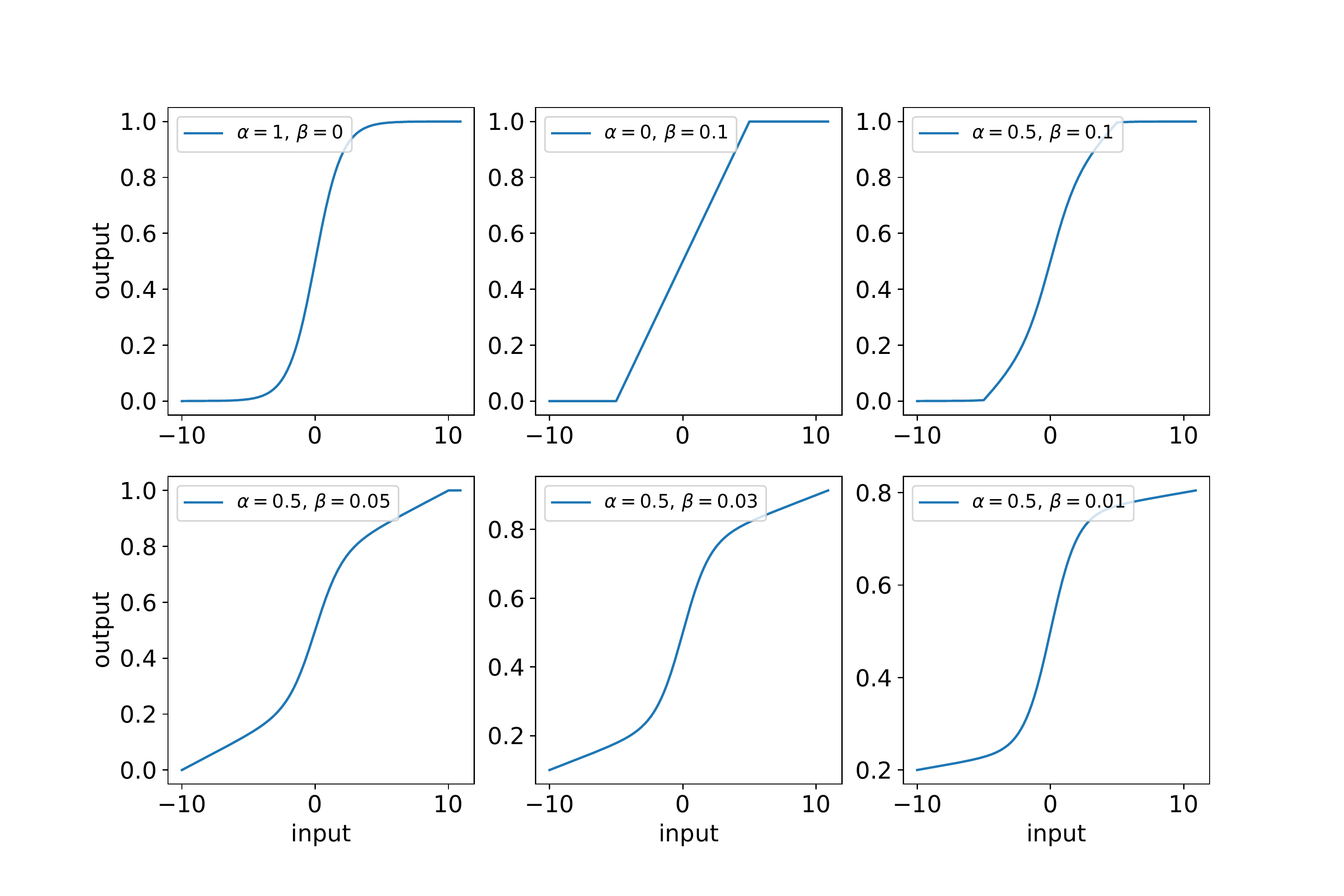}
 \caption{The shapes of combined activation function P-Sig-Ramp  proposed in Section ~\ref{Sec:2.2} with different set of parameters.}\label{Fig:1}
\end{center}
\end{figure}

Sigmoid and Tanh activation functions are widely used in recurrent neural networks, including basic recurrent nets and recurrent nets with cell structure such as LSTMs and GRUs. For the sigmoid function, the output should be in the domain of $[0,1]$, while for tanh the output should be in $[-1, 1]$. In the first case, one basic choice is:
\begin{equation}
    o(z;\alpha,\beta) = \alpha \cdot \sigma (z) + (1-\alpha) \cdot f(z; \boldsymbol{\beta})
    \label{Eq:4}
\end{equation}
where $0 \leq \alpha \leq 1$ and
\begin{equation}
 f(z; \boldsymbol{\beta}) =
  \begin{cases}
    0       & \quad \text{if } z < -\frac{1}{2\beta}\\
    \beta z+\frac{1}{2}  & \quad \text{if } -\frac{1}{2\beta} \leq z \leq \frac{1}{2\beta}\\
    1  & \quad \text{if } z > \frac{1}{2\beta}
  \end{cases}
  \label{Eq:5}
\end{equation}
In Eq.~\eqref{Eq:4}, $f(z; \beta)$ can be considered as a combination of two Ramp functions bounded between 0 and 1 with parameter $b$. 
The shapes of a sample of combined activation with Eq.~\eqref{Eq:4} are shown in Fig.~\ref{Fig:1}.
Similarly, we can build a function with the same boundary as $\tanh$ function, and use the corresponding combination to replace $\tanh$ in the LSTM cell. Consequently, for each combined flexible activation function, there are two parameters to be optimized during the training process. By combining the original activation function $\sigma$ with another function $f(z; \beta)$ with the same boundary using normalized weights, the model can be trained with flexible gates and have the potential to achieve better generalization performance. 

\subsection{P-E2-ReLU: ReLU Function substitute with unbounded domain} \label{Sec:2.3}
The outputs of ReLU function is unbounded on both sides, while the derivative with respect to the inputs is a step function. To build more flexible activation in the condition when ReLU function is used, we can make a weighted combination between ReLU and other non-linear functions with unbounded ends. Considering the goodness of candidate components as well as the principles proposed in Section ~\ref{Sec:2.1}, we introduce the following combined form:
\begin{equation}
\begin{split}
    o(z; \alpha, \beta) &= \alpha \text{ReLU}(z) + \beta \text{Elu}(z)\\
    &+ (1 - \alpha - \beta) (-\text{Elu}(-z))
\end{split}\label{Eq:6}
\end{equation}
This is actually a weighted combination of ReLU function and two ELU-family functions, which introduces non-linear functions with different shapes in both positive and negative region. If we bound all the wights between 0 and 1, each of the three components $\text{ReLU}(z)$, $\text{Elu}(z)$ or $-\text{Elu}(-z)$ can not be fully expressed by the combination of the others with certain model parameters. Figrue ~\ref{Fig:6} demonstrates the examples of P-E2-ReLU given by Equation ~\eqref{Eq:6} with different activation parameters.
\begin{figure}[th] 
\begin{center}
 \includegraphics[width=1.0\linewidth]{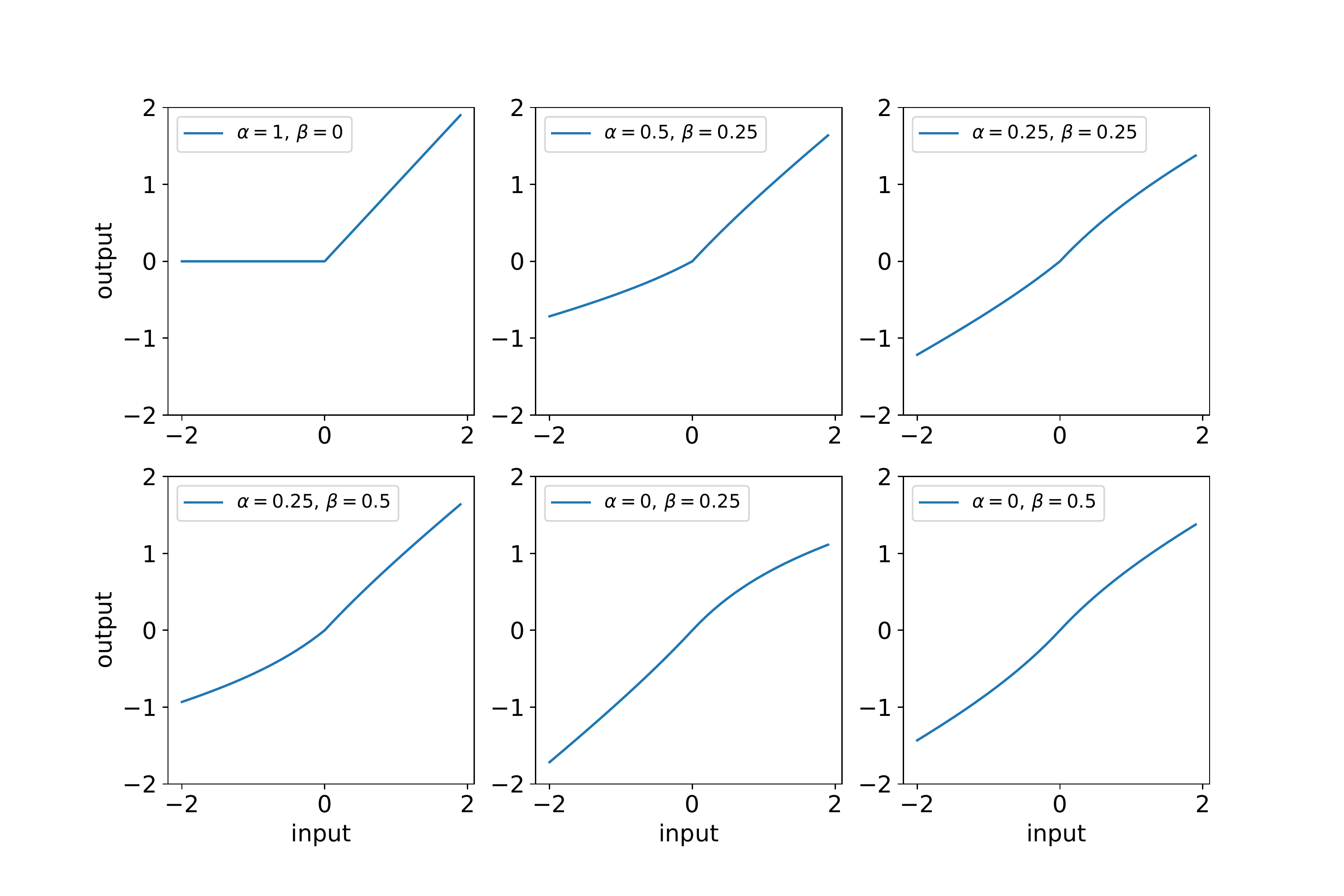}
 \caption{The shapes of combined activation function P-E2-ReLU proposed in Section ~\ref{Sec:2.3} with different set of parameters.}\label{Fig:2}
\end{center}
\end{figure} 
In addition, if we set $\beta = 1-\alpha-\beta = \frac{1-\alpha}{2}$, the combined two ELU-family terms will be symmetric with respect to point $(0,0)$, which is similar to the combination of $\tanh$ and ReLU proposed in \cite{manessi2018learning} but can be more naturally implemented for unbounded domains. Also, we can consider other combinations such as $\{\text{ReLU}, \text{Elu}(z)-\text{Elu}(-z)\}$, $\{\text{Id}, \text{Elu(z)}-\text{Elu(-z)}\}$ and $\{\text{ReLU}, \text{Elu}(z;\beta)-\text{Elu}(-z;\beta)\}$, where ``Id" represent identity functions, and $\beta$ is the trainable parameter of ELU (default is 1.0). We categorize them all as in ``P-E2-X" family. In this study we will focus on the form provided by Equation ~\eqref{Eq:6}, while the others could achieve even better performances in some tasks.

\subsection{Layer-wise Regularisation for Activation Parameters} \label{Sec:2.4}

Similar to the weights decay regularisation for model weights in NN models, we introduce regularisation terms for parameters in activation functions to avoid over-parameterization during learning process. When we set the summation of each component's weights in each flexible activation function to 1, it is not suitable to implement a L1 or L2 norm on the absolute value of activation weights. Instead, we use the L2 norm for the absolute difference between each specific activation parameter and the mean of corresponding parameters in the same layer. In addition, we introduce another L2 regularisation term controlling the difference between the trained parameters and the initial parameters of benchmark activation function. This can make sure that the benchmark is actually a specific case of flexible activation, while the variations can be learned to adapt to the training dataset and controlled by these regularisation effects. Thus, the cost function can be given by Equation ~\eqref{Eq:10a}, where $L_0$ is the predictive loss, $\alpha_{ijk}$ refers to the $k$th activation parameter $\alpha_k$ for $i$th element in $j$th layer, $\bar{\alpha}_{jk}$ is the average value of $\alpha_k$ in $j$th layer, $\alpha_{k0}$ is the combination coefficient of $k$th component in basic or standard activation functions (e.g. Sigmoid).
\begin{equation}
    \begin{split}
        & L = L_0 +\delta_1 \sum_{j}\frac{\lambda_j}{m_j}\sum_{i} \sum_{k} ||\alpha_{ijk}-\bar{\alpha}_{jk}||^2\\ 
        &+ \frac{\delta_2}{n}\sum_{i}\sum_{j}\sum_{k}||\alpha_{k0}-\alpha_{ijk}||^2\\ 
        & + \frac{\delta_3}{n} \sum_{i}\sum_{j}\sum_{k}(||\text{ReLU}(\alpha_{ijk^{*}}-(1-\Delta))||^2\\
        &+||\text{ReLU}(-\Delta
        -\alpha_{ijk^{*}})||^2) +\text{other terms}
    \end{split}\label{Eq:10a}
\end{equation}
 In the first regularization term, $m_j$ is the number of neurons in $j$th layer, while $\lambda_j$ is the layer-wise regularisation coefficients, and $\delta_1$ is mutual regularization coefficient. $\delta_2$ and $\delta_3$ are the regularization coefficients for the second and third regularization terms, and $n$ is the number of units in the whole network architecture. 
We can consider the first two regularization terms as prioris. For the first one, we consider that in the layer structure of deep neural networks, usually different layer is learning different level of patterns, which could be in favor of using similar activation functions in each layer. we call it as ``towards-mean regularization".   
Meanwhile, the second regularization terms can be considered as another priori in assuming that the initial activation functions are good enough and the learned activation parameters should not differ too much from the initial values. We call this as ``towards-default regularization". In addition, the third regularization term is introduced to ensure the combination parameters are bounded between 0 and 1, where $\Delta$ is a small number. We set $\delta_3$ to be a relatively large value (e.g. $\delta_3 = 1$) to make this as a strong constraint.
The regularization coefficients $\delta_1$ and $\delta_2$ can be implemented for controlling the flexibility in activation parameter space. The effects of these two regularization terms can be summarized as follows:
\begin{itemize}
    \item \textbf{Small} $\delta_1$, \textbf{small} $\delta_2$: The activation functions can be trained independently for different neurons in different layers with high flexibility.
    \item \textbf{Large} $\delta_1$, \textbf{small} $\delta_2$: The activation functions in the same layer will be constraint to be close to each other, while the activation in different layers can be different. 
    \item \textbf{Large} $\delta_2$: The activation function will be constraint to be close to the default fixed ones.
\end{itemize}
Therefore, with optimized activation coefficients, the models can achieve performances at least no worse than models with fixed default activation. Moreover, gradient-based methods may also be applied in optimizing these regularization coefficients \cite{maclaurin2015gradient, franceschi2017forward}.

\subsection{Initialization}
Since we introduced a set of new trainable parameters, the way to initialize them could make a difference. The default choice is to set the coefficients of benchmark activation functions to 1 and train the other terms from 0. However, this setting could suffer from the risk of being trapped in local optimal. Other solutions include the following:
\begin{itemize}
    \item Treat the initialization values as hyper-parameters, and use hyper-parameter searching algorithms to find the optimal setting. 
    \item Use random initialization following certain types of distributions with mode close 1 for the default activation.
\end{itemize}
The advantage of treating initialization values of activation parameters as hyper-parameters is that we can search for the optimal solution based on the final validation performance of model, while the drawback is higher computational cost for extra hyper-parameter optimizations. Alternatively, by using random initialization for each training, an ensemble of activation functions with different shapes are used at the beginning of model training, which can be more flexible in explore larger region in search space. In this study, to avoid tunning too many hyper-parameters, we initialized the parameters of activation function as standard activation functions such as sigmoid and ReLu, which corresponds to $\alpha=1$ in Eq.~\eqref{Eq:4} and Eq.~\eqref{Eq:6}.

\section{Experiments}\label{Sec:3}
All the experiments were conducted in the environment of Pytorch 1.3.1, we implemented the embedded functions of sigmoid, ReLU and pReLU in the baseline models and manually created the proposed flexible functions with backward path in the flexible models. The experiments are conducted with a cloud Intel Xeon 8-Core CPU.

\subsection{Experiment with Recurrent Neural Networks on multi-variate time series forecasting} \label{Sec:3.1}

For testing the performance of the model with flexible activation in recurrent neural networks, we build a multiple-layer LSTM model. We change the three sigmoid functions in Eq.~\eqref{Eq:3} to the parameterized combined function as shown in Eq.~\eqref{Eq:4}, then compare the model performances in the cases with or without flexible activations.

The datasets being experimented on is a combination of daily stock returns of G7 countries, which is a multi-variate time series \cite{selvin2017stock, fischer2018deep}. 
The returns of each day can be considered as an input vector to the corresponding hidden layer, while the output is one-step ahead forecast given a sequence of historical data. Instead of using random sampling, we directly split the set of sequences with 10 lagging vectors into training set (64\%), validation set (16\%) and test set (20\%), while the learning curves on training and validation sets can be obtained. The loss is selected as the average of mean squared errors of all forecasted values with respect to the true values for each example. For the hyper-parameter setting, the batch size was set to be 50, the window size is 10 time steps as is recommended by related papers\cite{qin2017dual}, while the optimizer implemented in training is Adam optimizer with the same learning rate of 0.001 on both weights, bias and activation parameters. The initialization of the flexible activation parameters in replacing sigmoid function is $\alpha = 1$ and $\beta = 0.1$, which means that we train them from baseline settings. On the other hand, four stacked LSTM models with different layer configurations are implemented, then we compare the validation and test performances of these models with (a) fixed activations, (b) flexible activations without regularization, (c) flexible activation with a toward-mean regularization using a same default regularization coefficient of 0.025. We perform 50 trials for each setting with different random initialization for each of them.

\begin{figure}[htb] 
\begin{center}
 \includegraphics[width=1.0\linewidth]{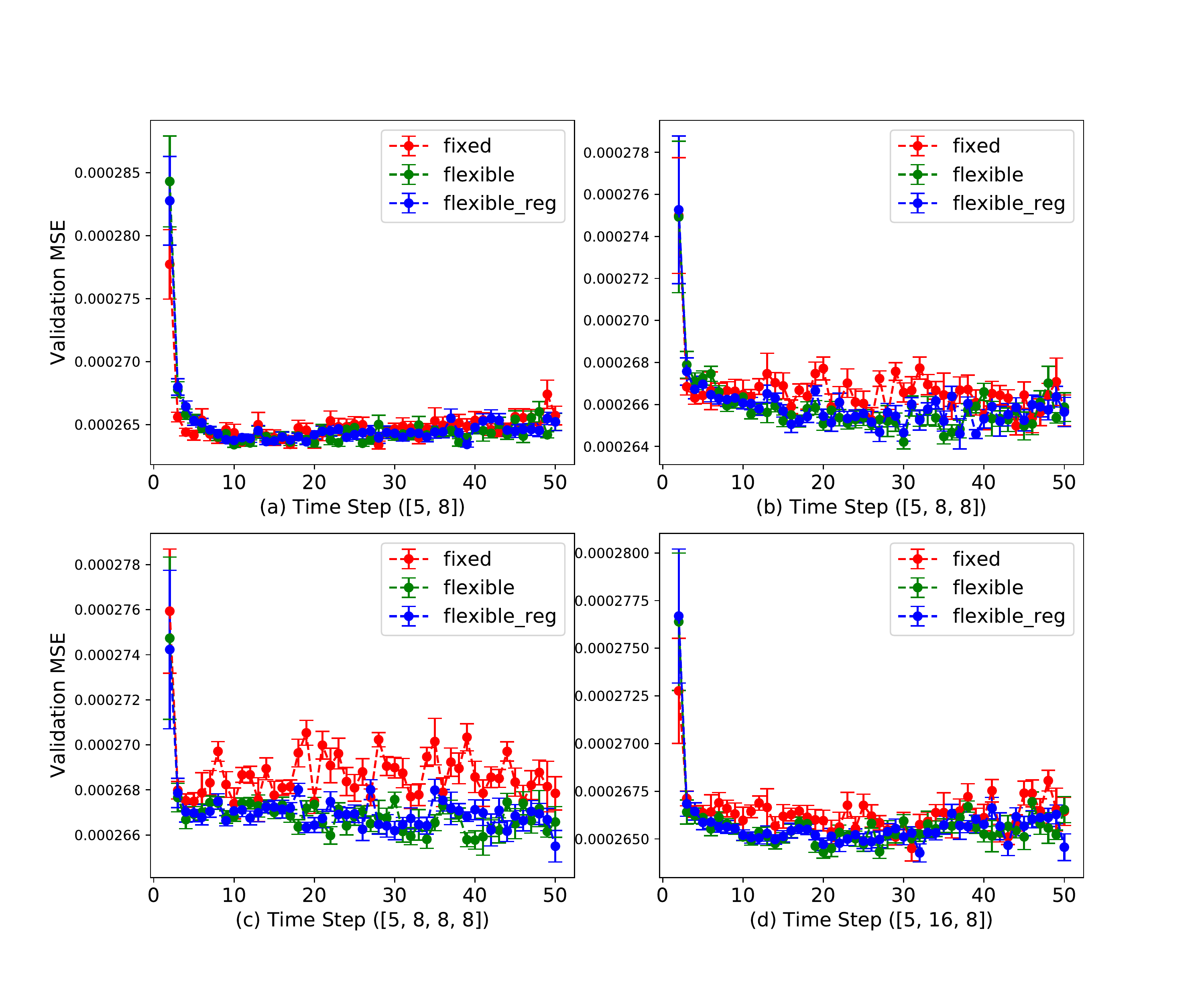}
 \caption{Comparison between the average learning curves (with error bars) of LSTM models with and without regularized flexible activation functions on  forecasting BRICKS Indice returns. (a) Layer size: $[5, 8]$; (b) Layer size: $[5, 8, 8]$; (c) Layer size: $[5, 8, 8, 8]$; (d) Layer size: $[5, 16, 8]$.}\label{Fig:3.3}
\end{center}
\end{figure}

\begin{figure}[th] 
\begin{center}
\vspace{-0.3cm}
\setlength{\belowcaptionskip}{-5pt}
 \includegraphics[width=1.0\linewidth]{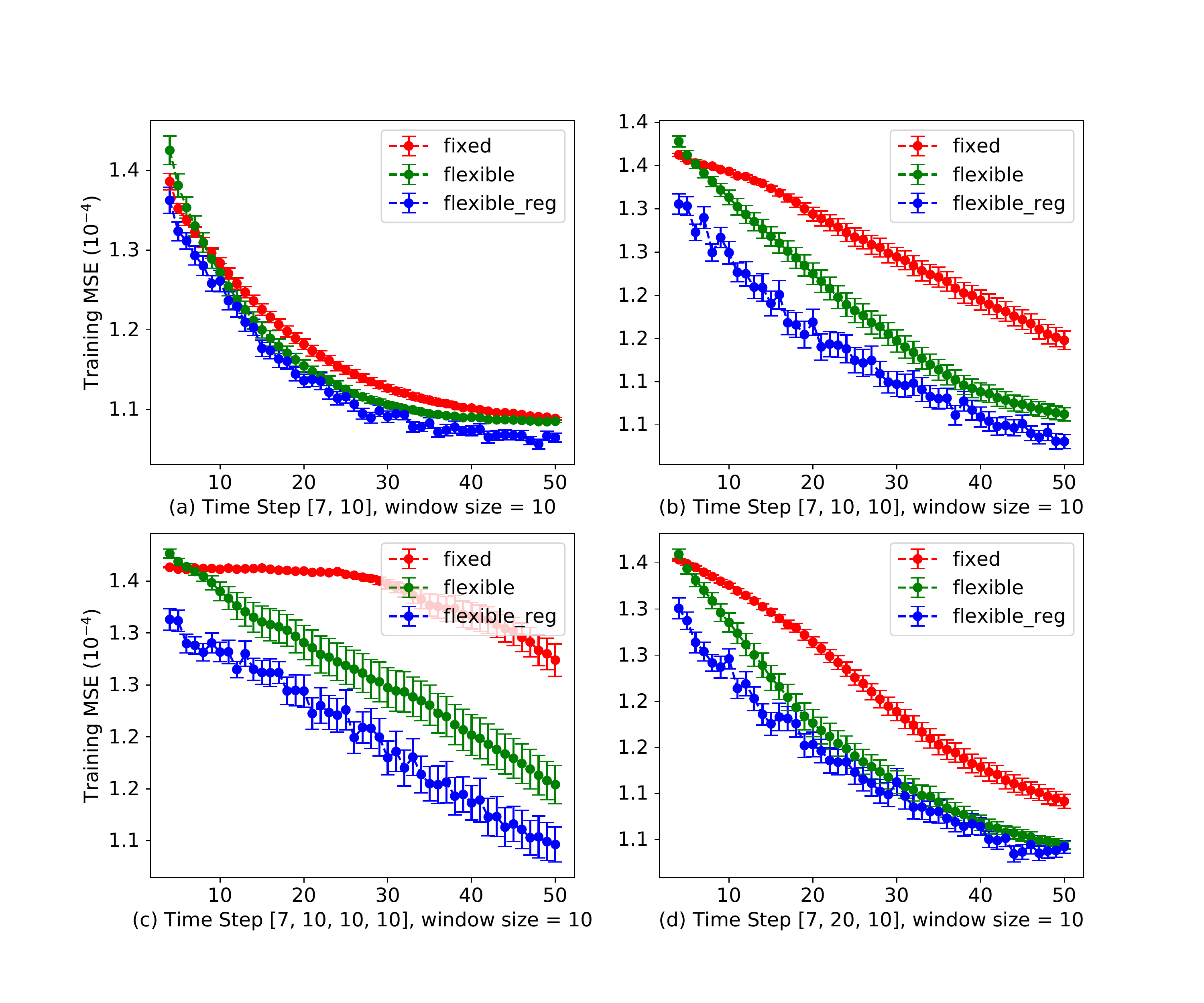}
 \caption{Comparison between the average training curves (with error bars) of LSTM models with and without regularized flexible activation functions on  forecasting G7 Indice returns. (a) Layer size: $[7, 10]$; (b) Layer size: $[7, 10, 10]$; (c) Layer size: $[7, 10, 10, 10]$; (d) Layer size: $[7, 20, 10]$.}\label{Fig:3.4a}
\end{center}
\end{figure}

\begin{figure}[!htb] 
\begin{center}
 \includegraphics[width=1.0\linewidth]{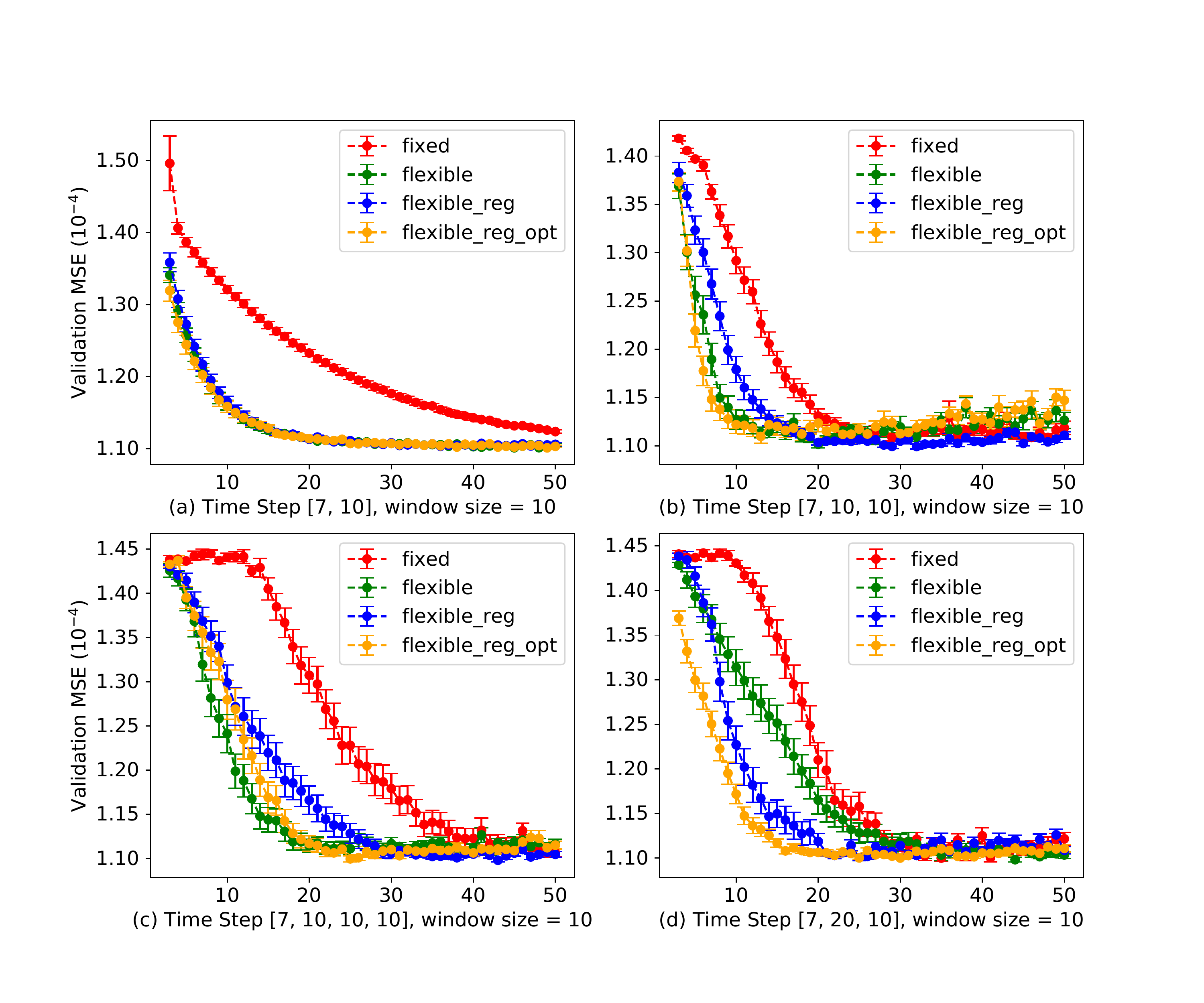}
 \caption{Comparison between the average learning curves (with error bars) of LSTM models with and without regularized flexible activation functions on  forecasting G7 Indice returns. (a) Layer sizes: $[7, 10]$; (b) Layer sizes: $[7, 20]$; (c) Layer sizes: $[7, 10, 10, 10]$; (d) Layer sizes: $[7, 20, 10]$.}\label{Fig:3.5}
\end{center}
\end{figure}

For BRICS indices forecasting, we test the model performances with 4 stacked LSTM configurations, including models with layer structures: $[5, 8]$, $[5, 8, 8]$, $[5, 8, 8, 8]$ and $[5, 16, 8]$. For example, in the first layer structure, we have 5 input units for each time step, and a LSTM layer with a hidden state of length 8 for each time step. The results are provided in Fig.~\ref{Fig:3.3}, where we use error bars to show the sample standard errors of each method in each time step.
As we can see, for all the four cases, generally the learning curves of models with flexible activation and regularized flexible activation (with fixed towards-mean regularization) lie below the learning curves of models with fixed sigmoid activations. For G7 indices forecasting, we exam the performances on stacked-LSTM models with layer structures: $[7, 10]$, $[7, 20]$, $[7, 10, 10, 10]$ and $[7, 20, 10]$. The results on training set are provided in Fig.~\ref{Fig:3.4a}. The curves are plotted from the 5th epoch to focus on the comparison in later stage. 
We can see that in Fig.~\ref{Fig:3.4a}, for all the four cases, the learning curves of model with flexible activation functions (denoted as ``flexible models'') on training set lie below the corresponding curves of models with fixed activation functions (denoted as ``fixed models'') during most of training time. In addition, towards-mean regularization on activation parameters can further improve the convergence performance. Meanwhile, the error bars of population means demonstrate that the improvement of flexible models in terms of validation performance is significant in general. 
\begin{figure}[htb] 
\begin{center}
 \includegraphics[width=1.0\linewidth]{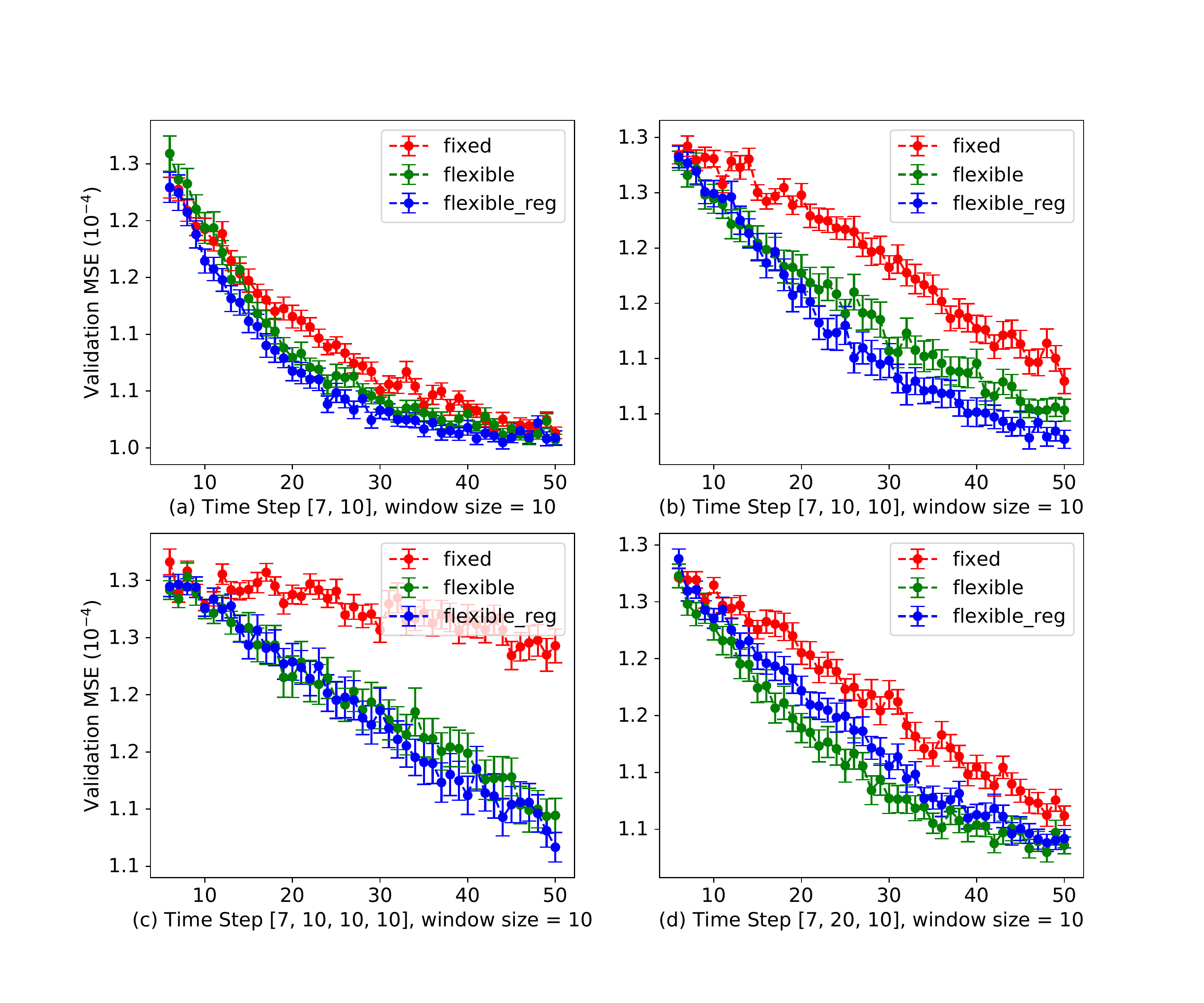}
 \caption{Comparison between the average learning curves (with error bars) of LSTM models with and without regularized flexible activation functions on  forecasting G7 Indice returns. (a) Layer size: $[7, 10]$; (b) Layer size: $[7, 10, 10, 10]$; (c) Layer size: $[7, 20, 10]$; (d) Layer size: $[7, 20, 10]$.}\label{Fig:3.4}
\end{center}
\end{figure}
Similarly, we can see that in Fig.~\ref{Fig:3.4}, for all the four cases, the learning curves of flexible models on validation set lie below the corresponding curves of fixed models during most of training time. Moreover, the error bars of population means demonstrate that the improvement of flexible models in terms of validation performance is significant in general. To make a further investigation, we use optimized learning rates instead of the default ones in each case. First, we perform grid searches for both models without flexible activation functions and model with flexible activation functions in the range of $[0.001, 0.1]$ on logarithm scale. Also, for models with flexible activation, we perform another grid search in the range of $[0.01, 0.2]$ on logarithm scale for the optimal activation regularization coefficients defined in Section ~\ref{Sec:2.4}, while the searching is based on the optimal learning rate in each case. Then for each architecture, we compare the models with four settings: (a) Models without flexible activation on the corresponding optimal learning rate; (b) Models with flexible activation function on the corresponding optimal learning rate, without regularization on the activation function; (c) Models with flexible activation function on the corresponding optimal learning rate, with regularization on the activation function on the default value of 0.025; (d) Models with flexible activation function on the corresponding optimal learning rate, with optimal regularization on the activation function in each case. The results of 50 runs with corresponding error bars are shown in Fig.~\ref{Fig:3.5}.
We can learn from Fig.~\ref{Fig:3.5} that with optimized learning rates and Adam optimizer, LSTM models with flexible activation functions perform better in terms of convergence compared with fixed models for all the four architectures. By introducing regularization on flexible activation towards the mean, the convergence can be further improved. To investigate the effect of two types of regularization, we conduct a simulation and plot the relationship between the minimum validation model performance and the values of regularization coefficients. We use a default learning rate of 0.001. For regularization coefficients, we take 20 discrete grid points in the logarithm scale of 0.001 to 0.2. At each point, we plot the average of 30 trials with corresponding error bars for the standard error of sample means.
\begin{figure}[th]
\begin{center}
 \includegraphics[width=1.0\linewidth]{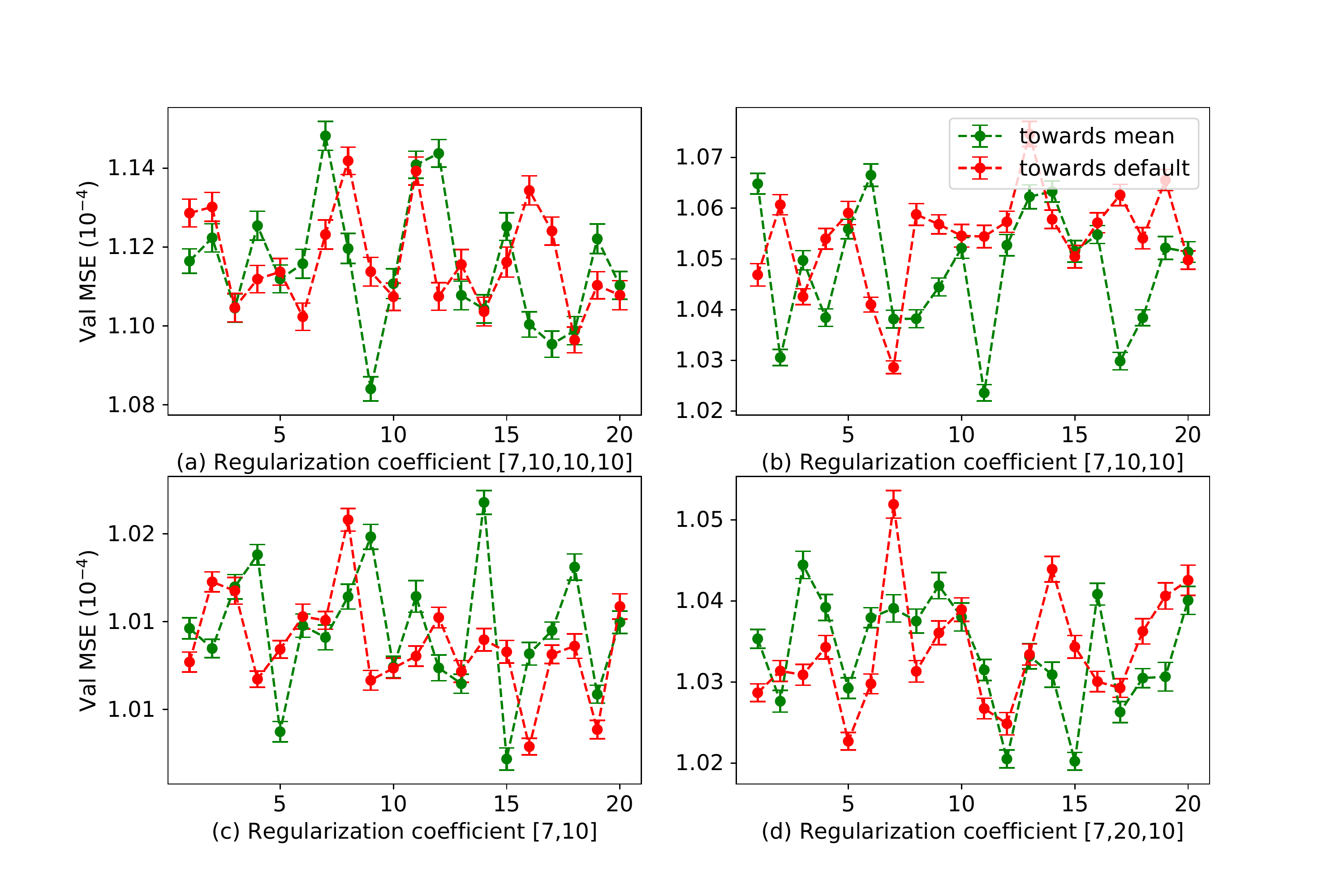}
 \caption{Comparison between the towards-mean and towards-default regularizations with different regularization coefficients on minimum validation MSE. The architectures of four subplots are same with those in Fig.~\ref{Fig:3.5}.
 }\label{Fig:6}.
\end{center}
\end{figure}
It is shown in Fig.~\ref{Fig:6}, the minimum validation model performance is not constant under different regularization coefficients for both the two types of regularization of flexible activation functions. By referring to the error bars for 30 trials, we can learn that the change is not random, but with statistical significance. Generally, the minimum validation performances are obtained by regularization towards the mean with a intermediate regularization coefficient. Also, we have had a similar investigation based on the minimum validation loss in each setting. We use the test set of G7 for testing and all training and validation samples for training, with a default learning rate of 0.001. At each point, we plot the average of 20 trials with corresponding error bars for the standard error of sample means.
\begin{figure}[th]
\begin{center}
 \includegraphics[width=1.0\linewidth]{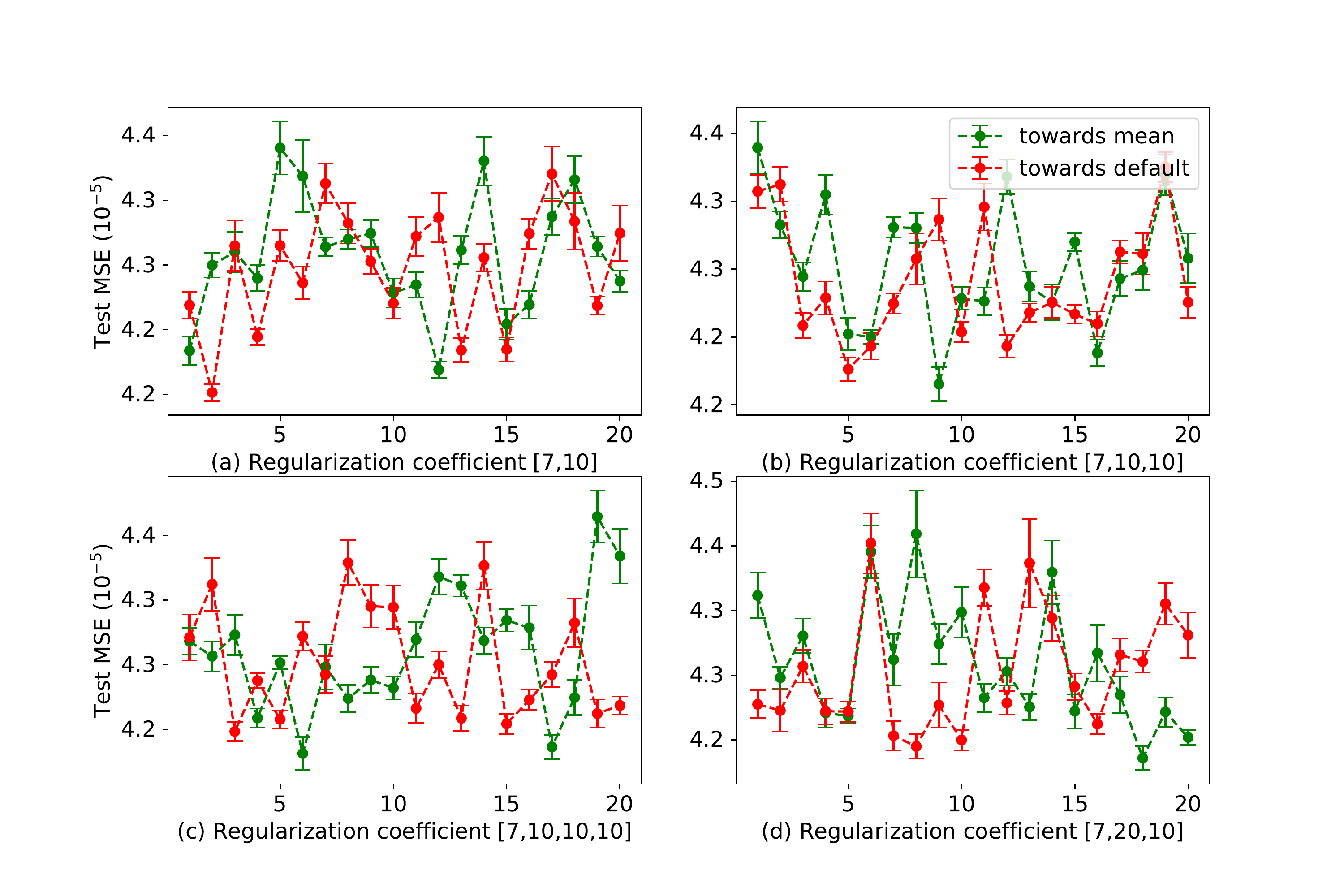}
 \caption{Comparison between the towards-mean and towards-default regularizations with different regularization coefficients on test set. The architectures of four subplots are same with those in Fig.~\ref{Fig:3.5}.
 }\label{Fig:6a}.
\end{center}
\end{figure}
As is shown in Fig.~\ref{Fig:6a}, the test model performance is not constant under different regularization coefficients for both the two types of regularization of flexible activation functions. By referring to the error bars for 20 trials, we can learn that the change of test performance is not random, but with statistical significance. Generally, the best performances are obtained with a intermediate regularization coefficient. This demonstrates that a better test performance can be achieved by optimizing the towards-mean or towards default regularization of flexible activation function given the architectures unchanged. 

\subsection{Experiment with convolutional autoencoder} \label{Sec:3.2}
\begin{figure}[th] 
\begin{center}
 \includegraphics[width=1.0\linewidth]{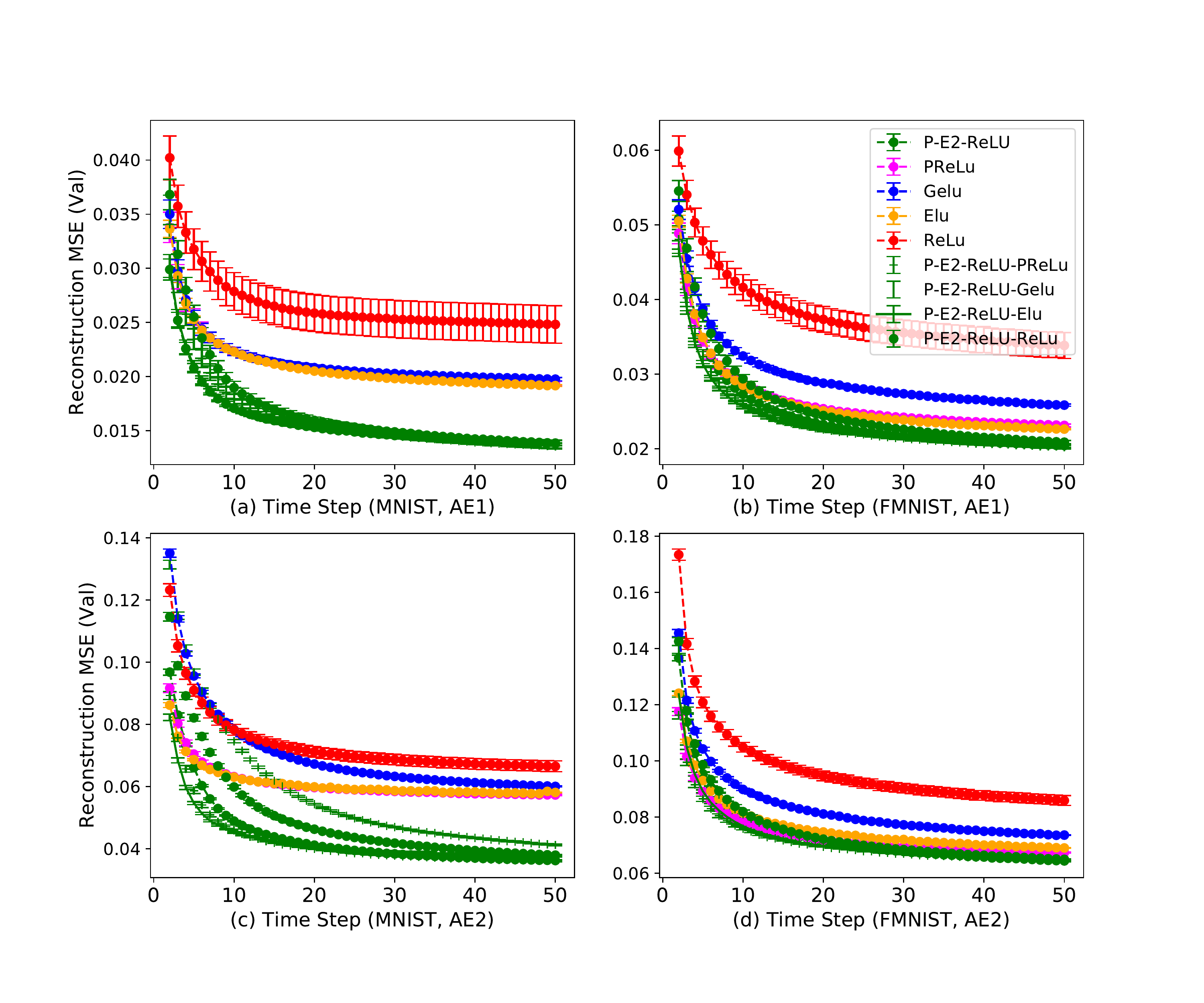}
 \caption{Comparison between the average learning curves (with error bars) of convolutional auto-encoder models with different activation functions on image compression task. (a) CAE 1 model for MNIST; (b) CAE 1 model for FMNIST; (c) CAE 2 model for MNIST; (d) CAE 2 model for FMNIST.}\label{Fig:7}
\end{center}
\end{figure}

\begin{figure}[th] 
\begin{center}
 \includegraphics[width=1.0\linewidth]{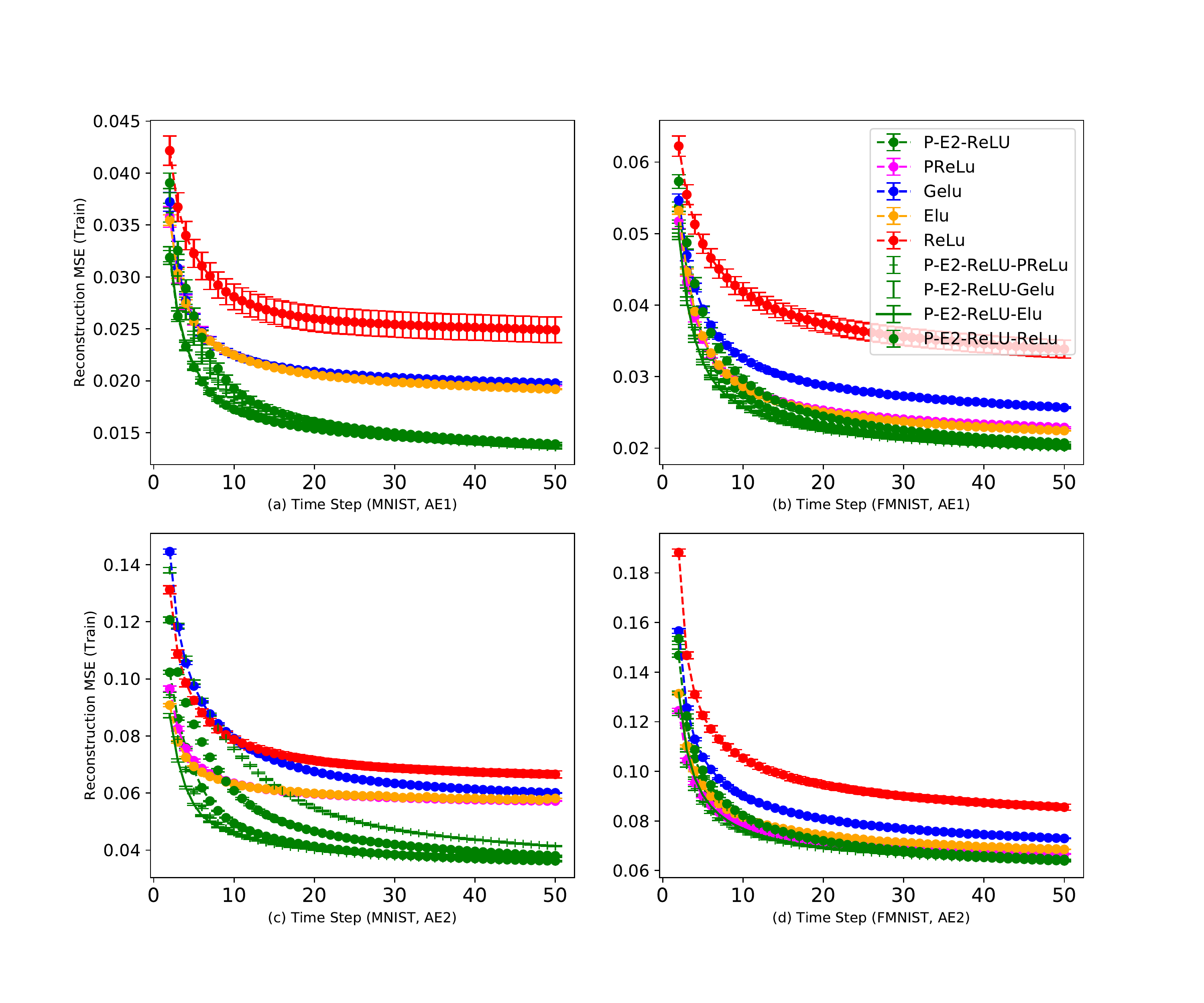}
 \caption{Comparison between the average learning curves (with error bars) of convolutional auto-encoder models with different activation functions on image compression task. (a) CAE 1 model for MNIST; (b) CAE 1 model for FMNIST; (c) CAE 2 model for MNIST; (d) CAE 2 model for FMNIST.}\label{Fig:7a}
\end{center}
\end{figure}

The second experiment is conducted with autoencoder models for lossy image compression on several benchmark datasets including MNIST and FMNIST \cite{goodfellow2016deep}. The architecture used in this experiment can be shown as follows: 
\newline
\textbf{Convolutional Autoencoder 1 (CAE 1)}:
\begin{equation*}
\begin{split}
&\text{Input(28*28*1)} \rightarrow \text{Conv2d(16, 3, 3)} \xrightarrow{\text{ReLU}} \text{MP(2, 2)}\rightarrow \\
& \text{Coding} \rightarrow \text{Conv2d(8, 5, 3)} \xrightarrow{\text{ReLU}} \text{Conv2d(1, 2, 2 )}\\
&\xrightarrow{\text{Tanh}} \text{Output} 
\label{Eq:9}
\end{split}
\end{equation*}
\textbf{Convolutional Autoencoder 2 (CAE 2)}:
\begin{equation*}
\begin{split}
&\text{Input(28*28*1)} \rightarrow \text{Conv2d(16, 3, 3)} \xrightarrow{\text{ReLU}} \text{MP(2, 2)}\rightarrow\\
&\text{Conv2d(8, 3, 2)}\xrightarrow{\text{ReLU}} 
\text{MP(2, 1)}\rightarrow
\text{Coding}\\ 
&\rightarrow \text{Conv2d(16, 3, 2)} \xrightarrow{\text{ReLU}} \text{Conv2d(8, 5, 3)}\\
&\xrightarrow{\text{ReLU}}
\text{Conv2d(1, 2, 2 )}\xrightarrow{\text{Tanh}} \text{Output} 
\label{Eq:9a}
\end{split}
\end{equation*}
\textbf{Convolutional Autoencoder 3 (CAE 3)}:
\begin{equation*}
\begin{split}
&\text{Input(32*32*3)} \rightarrow \text{Conv2d(12, 4, 2)} \xrightarrow{\text{ReLU}} \text{Conv(24, 4, 2)}\\ 
&\xrightarrow{\text{ReLU}}\text{Conv(48, 4, 2)}\xrightarrow{\text{ReLU}}\text{Coding}
\rightarrow \text{Conv2d(24, 4, 2)}\\
&\xrightarrow{\text{ReLU}} \text{Conv2d(12, 4, 2)}\xrightarrow{\text{ReLU}}\text{Conv2d(3, 4, 2)}\xrightarrow{\text{ReLU}}\text{Output} 
\label{Eq:10}
\end{split}
\end{equation*}
CAE 1 and CAE 2 compress images of size 28*28 to 16 filters with size 2*2 for each. For each trial, we randomly sample 5,000 examples from the original training datasets of MNIST and FMNIST as the training data, another randomly sampled example from the remaining of the training set as the validation set, and use the original test sets as the test sets in our experiment. We then compare the performances of following activation functions in replacing the ReLU activation functions in the original architectures of CAE 1 and CAE 2:
(a) ReLU function; (b) PReLU function; (c) ELU function; (d) Gelu function; (e) Newly proposed P-E2-ReLU function, which is initialized as $o(z) = 0.4\text{ReLU}+0.3\text{Elu}(z)+0.3(-\text{Elu}(-z))$. The learning rate of each model with different activation function is optimized with a grid-search in the range between 0.001 and 0.1 on logarithmic scale. In addition, we run four more models in each trials with P-E2-ReLU on the optimized learning rates for other four candidates activation functions. After 50 trials, we can compare the learning curves of these five activation functions. The validation curves are given in Fig.~\ref{Fig:7}, while the training curves are given in Fig.~\ref{Fig:7a}. 

\begin{figure}[th]
\begin{center}
 \includegraphics[width=1.0\linewidth]{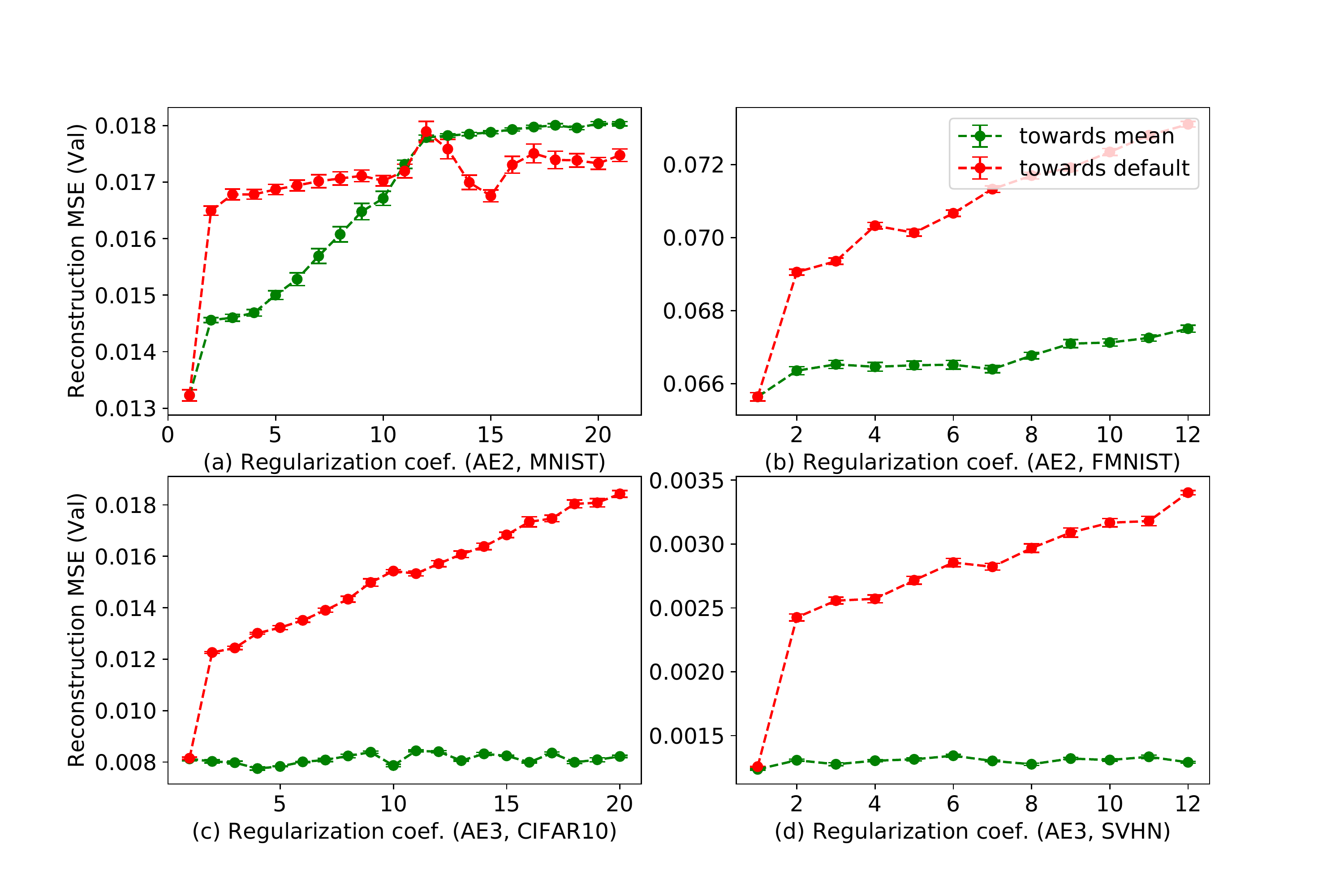}
 \caption{Comparison between the average learning curves (with error bars) of Autoencoder models with flexible activation P-E2-ReLU under different activation regularization settings.}\label{Fig:9}.
\end{center}
\end{figure}

We noticed that models with the newly proposed P-E2-Relu activation function outperform all the other activation function in terms of the both validation and training performance with statistical significance during the whole learning process after a small amount of initial steps. Meanwhile, the standard ReLU activation function perform the worst, and the other three activation functions including PReLU, ELU and Gelu get similar performances in general. Moreover, even with the learning rates optimized under models with other activation functions, the proposed P-E2-ReLU activation function still outperform other activation functions in this task. To make a further investigation, we perform experiments on CIFAR10 and SVHN with another convolutional auto-encoder architecture given by CAE 3, where the encoder compress each image of size 32*32*3 to 48 filters with size 4*4. In addition, we introduced other two flexible activation functions proposed in Section ~\ref{Sec:2.3}, which are combinations of $\{\text{ReLU}, \text{Elu}(z)-\text{Elu}(-z)\}$ and $\{\text{Id}, \text{Elu(z)}-\text{Elu(-z)}\}$. The initialization of the two combination weights for these two activation functions are both 0.5. The learning rate is set to be the default value of 0.001 for all activation functions.
\begin{figure}[th] 
\begin{center}
 \includegraphics[width=1.0\linewidth]{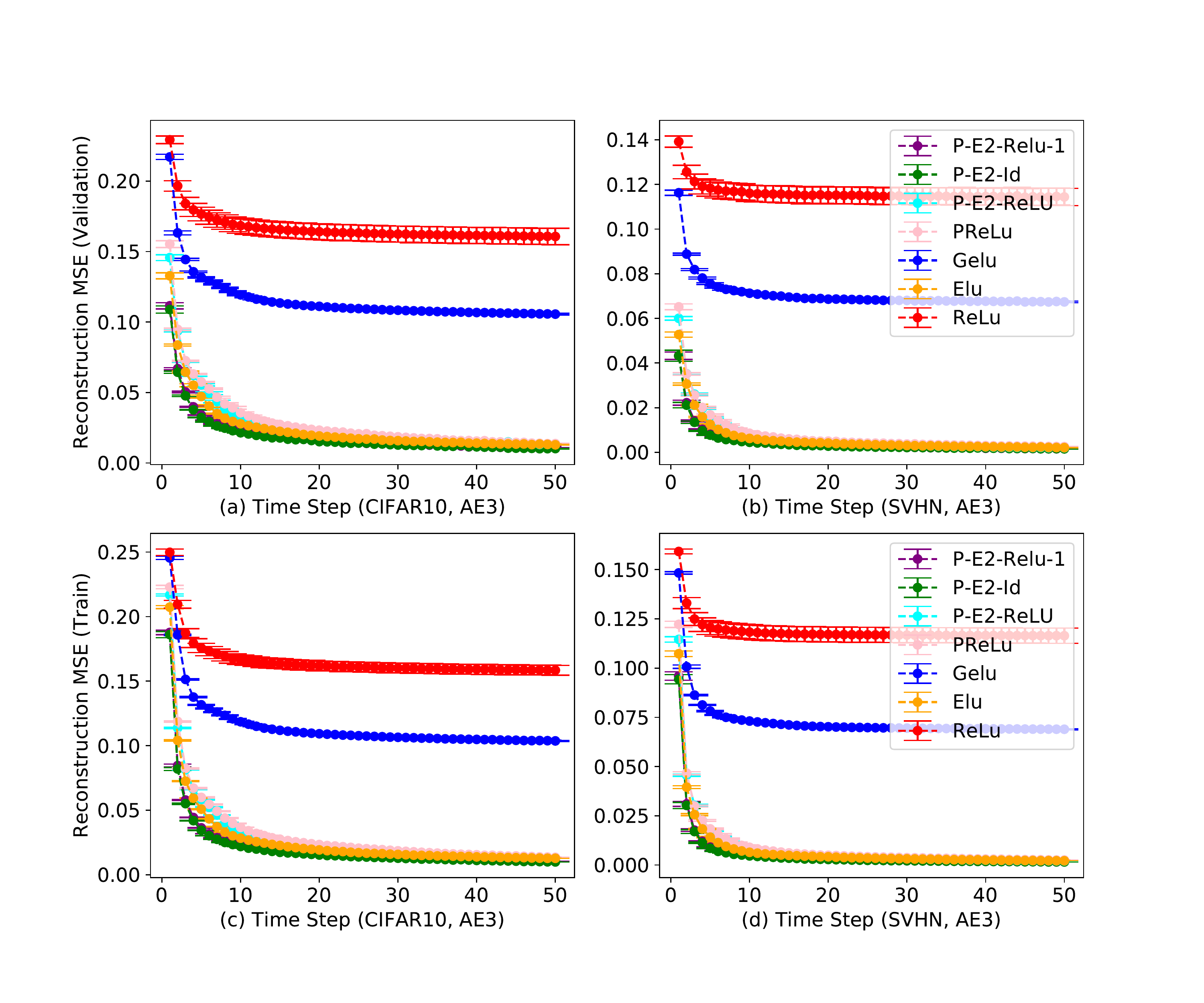}
 \caption{Comparison between the average learning curves of convolutional auto-encoder models with different activation functions on image compression task. (a) CAE 3 model for CIFAR10; (b) CAE 3 model for SVHN.}\label{Fig:8}
\end{center}
\end{figure}
As is shown in Fig.~\ref{Fig:8}, the experiment with CAE 3 architecture show the advantage of bounded P-E2-ReLU and P-E2-Id activation functions over other state-of-the-art activation functions in the image compression tasks on CIFAR10 and SVHN. For both two datasets, the model with the proposed P-E2-Id activation function perform the best, although the performance is close to ELU and PReLU. Further, we made one-tail two sample t-test for the difference of population means on test set. The results are given in Table ~\ref{tab:1}, where the null hypothesis $\text{H0: } m_4\geq m_2$ means the population mean test loss of bounded P-E2-ReLU ($m_4$) is worse or equal to that of ELU ($m_2)$. The test statistics demonstrate that P-E2-ReLU significantly outperforms PReLU and ELU in terms of test reconstruction MSE in both cases, with very small p-values, while P-E2-Id performs even better. 
\begin{table}[htbp]
  \centering
  \caption{Statistical Tests for Performances on Test Set}
    \begin{tabular}{cccc}
    \toprule
    \multirow{2}[2]{*}{} & \multirow{2}[2]{*}{Activation} & \multicolumn{2}{c}{Dataset} \\
          &       & CIFAR10 & SVHN \\
    \midrule
    Model 1 & P-E2-Id & \multicolumn{1}{l}{\textbf{1.01E-2 (2.4E-4)}} & \textbf{1.25E-3 (3.9E-5)} \\
    Model 2 & ELU   & \multicolumn{1}{l}{1.27E-2 (1.6E-4)} & 1.84E-3 (2.7E-5) \\
    Model 3 & PReLU & \multicolumn{1}{l}{1.35E-2 (1.5E-4)} & 2.24E-3 (9.3E-5) \\
    Model 4 & P-E2-ReLU & \multicolumn{1}{l}{1.05E-2 (2.8E-4)} & 1.38E-3 (6.3E-5) \\
    \midrule
          & Null Hypothesis & \multicolumn{2}{c}{p-value} \\
    \midrule
    Test 1 &  H0: $m_1\geq m_2$     & 3.64E-09 & 3.96E-14 \\
    Test 2 &  H0: $m_1\geq m_3$     & 1.73E-13 & 3.97E-11 \\
    Test 3 &  H0: $m_4\geq m_2$     & 2.05E-06 & 4.86E-7 \\
    Test 4 &  H0: $m_4\geq m_3$     & 4.77E-10 & 2.03E-7 \\
    \bottomrule
    \end{tabular}%
  \label{tab:1}%
\end{table}%
Again, to check the effect of regularization on flexible activations, we train and evaluate flexible models under different regularization coefficients. The learning rate is optimized for P-E2-ReLU. We take 20 discrete grid points from 0.001 to 0.2 in logarithm scale. The result is shown in Fig.~\ref{Fig:9}.
As we can see in Fig.~\ref{Fig:9}, for towards-default regularization, the models achieve the best performances under the smallest regularization coefficients.
This indicates that model performance can be significantly improved by introducing the proposed flexible activating function. On the other hand, for towards-mean regularization, in half of the cases the minimum validation errors are obtained when the regularization coefficients are 0, while in the other two cases, the minimum validation errors are obtained at intermediate points. In addition, we notice that the curves representing towards-mean regularization are below the curves of towards-default regularization in most of the time. In (c) and (d), a strong towards-mean regularization does not harm the model performance but a strong towards-default regularization does, which indicates that the models can be improved by sharing different activations functions in different layers without variation across each same layer, and we may only need two more parameters in each layer to achieve this improvement with layer-wise learned activation functions.

\section{Number of model parameters and space complexity}\label{Sec:4}

\subsection{Number of parameters and model performances}\label{Sec:4.1}
When batch-normalization is not implemented, the number of parameters in each deep neural network is equal to the number of weighs plus the number of bias. For an FFNN, $N = N_W + N_b =\Sigma_{i=1}^{L-1}n_i n_{i+1}+\Sigma_{i=2}^{L}n_i$, where $L$ is the number of layers including input layer, hidden layer and output layer, while $n_i$ is the number of units in $i$th layer. If flexible activations are introduced in all hidden and output units, with two independent extra parameters in each unit, the total number of model parameters will increase by $N_a = 2\sum_{i=2}^{L}n_i$.  The number of parameters in basic RNN models is depend on the type of hidden cells used, in general it is $N = g\sum_{i=1}^{L-1} (n_{i+1}(n_{i+1}+n_i) +n_{i+1})$, where $g$ is the number of weight matrices in each cell. For basic RNN, $g=1$, for GRU, $g=3$, and for LSTM, $g=4$. Meanwhile, the extra number of parameters needed for RNNs with flexible activations is $N_a = 2\sum_{i=2}^{K}s_i n_i$, where $s_i$ is the number of activation functions replaced by flexible ones in $i$th layer.

Following this derivation, in the LSTM models used in Section~\ref{Sec:3.1}, we can make a summary table for the increasing ratio of model parameters in each case. 
\begin{table}[htbp]
  \centering
  \caption{Comparison of number of parameters in different settings of LSTM models.}
    \begin{tabular}{ccccc}
    \toprule
    \multicolumn{1}{p{4 em}}{model} & \multicolumn{1}{p{4 em}}{layer size} & \multicolumn{1}{p{4 em}}{basic model} & \multicolumn{1}{p{4 em}}{flexible activation} & \multicolumn{1}{p{4 em}}{ratio of increase} \\
    \midrule
    1     & [5, 8] & 448  & 48   & 10.7\% \\ 
    2     & [5, 8, 8] & 992   & 96    & 9.6\% \\ 
    3     & [5, 8, 8, 8] & 1536  & 144    & 9.3\% \\ 
    4     & [5, 16, 8] & 2208  & 144   & 6.5\% \\ 
    \bottomrule
    \end{tabular}%
  \label{tab:3.1}%
\end{table}%
\begin{table}[htbp]
  \centering
  \caption{Comparison of number of parameters in different settings of LSTM models for G7 indices forecasting.}
    \begin{tabular}{ccccc}
    \toprule
    \multicolumn{1}{p{4 em}}{model} & \multicolumn{1}{p{4 em}}{layer size} & \multicolumn{1}{p{4 em}}{basic model} & \multicolumn{1}{p{4 em}}{flexible activation} & \multicolumn{1}{p{4 em}}{ratio of increase} \\
    \midrule
    1     & [7, 10] & 720  & 60  & 8.3\% \\ 
    2     & [7, 10, 10] & 1560   & 120    & 7.6\% \\ 
    3     & [7, 10, 10, 10] & 2400  & 180    & 7.5\% \\ 
    4     & [7, 20, 10] & 3480  & 180   & 5.1\% \\ 
    \bottomrule
    \end{tabular}
  \label{tab:3.1a}%
\end{table}%
As shown in Table~\ref{tab:3.1} and Table~\ref{tab:3.1a}, we can see that the models with flexible activations have about 5\% to 10\% increase in the number of parameters. However, it is shown in Section ~\ref{Sec:3.1} that model 1, model 3 and model 4 have very similar validation performances during training even though their number of parameters varies in much higher (even more than 100\%) proportions, while the corresponding validation performances for models with flexible activations are better than all the models with fixed activations. 

In Table~\ref{tab:3.1}, the best performed model with layer size $[5, 8]$ has a single hidden layer and the number of parameters is smaller than the other models with layer sizes $[5,8,8]$, $[5, 16, 8]$ and $[5,8,8,8]$. Similarly, in Table~\ref{tab:3.1a}, the best performed model with layer size $[7, 10]$ has a single hidden layer and the number of parameters is smaller than the other models with layer sizes $[7,10,10]$, $[7, 20, 10]$ and $[7,10,10,10]$. Therefore, the number of parameters does not really matters in this range when we consider the minimum validation performance in a long training time, and the improvement of model performance can be explained by the advantage of introducing flexible activation with trainable parameters.
\begin{table}[htbp]
  \centering
  \caption{Comparison of number of parameters in different settings of CAE models.}
    \begin{tabular}{ccccc}
    \toprule
    \multicolumn{1}{p{4 em}}{model} & \multicolumn{1}{p{4 em}}{layer size} & \multicolumn{1}{p{4 em}}{basic model} & \multicolumn{1}{p{4 em}}{flexible activation} & \multicolumn{1}{p{4 em}}{ratio of increase} \\
    \midrule
    1     & CAE 1 & 3401  & 50   & 1.47\% \\ 
    2     & CAE 2 & 5729  & 66    & 1.15\% \\ 
    3     & CAE 3 & 47355  & 168  & 0.35\% \\ 
    \bottomrule
    \end{tabular}%
  \label{tab:3.2}%
\end{table}%
\begin{table}[htbp]
  \centering
  \caption{Comparison of number of parameters in different settings of CAE models with P-E2-ReLU-1/P-E2-Id.}
    \begin{tabular}{ccccc}
    \toprule
    \multicolumn{1}{p{4 em}}{model} & \multicolumn{1}{p{4 em}}{layer size} & \multicolumn{1}{p{4 em}}{basic model} & \multicolumn{1}{p{4 em}}{flexible activation} & \multicolumn{1}{p{4 em}}{ratio of increase} \\
    \midrule
    1     & CAE 1 & 3401  & 25   & 0.73\% \\ 
    2     & CAE 2 & 5729  & 33    & 0.57\% \\ 
    3     & CAE 3 & 47355  & 84  & 0.17\% \\ 
    \bottomrule
    \end{tabular}%
  \label{tab:3.2a}%
\end{table}%
For the three Convolutional auto-encoder models in Section~\ref{Sec:3.2}, the basic model parameters include those in convolutional layers. The number of parameters of each convolutional neural layer is: $N_c = (n*n*l+1)*k$, where $n$ is the filter size, while $l$ and $k$ represent the numbers of input and output channels. For filter-shared flexible activation functions with $p$ activation parameters, the extra number of parameters is $N_a = p*k$, where $p=1$ for P-ReLU-Id and P-ReLU-E2-1, and $p=2$ for P-ReLU-E2. The summary table showing the numbers of basic model parameters as well as extra activation parameters for three CAE models are given in Table ~\ref{tab:3.2} and ~\ref{tab:3.2a}. We can see that although the experimental results shown in Fig.~\ref{Fig:7a} and Fig.~\ref{Fig:8} demonstrate that the proposed P-E2-ReLU achieves a significant improvement compared with fixed activation such as ReLU and GeLU, the extra number of parameter introduced only a quite small proportion of the total number of parameters in each architecture. Moreover, as is indicated in Fig.~\ref{Fig:9}, when layer-wise flexible activation are applied, the model performance can be improved with even smaller number of extra activation parameters.

To sum up, the results at least show that we can achieve a quite significant performance improvement with only a small proportion of extra parameters in the flexible activation functions. Although the same mapping may be learned without flexible activation functions, it may need much more number of parameters or much larger effort of hyper-parameter searching with a bunch of different model architectures.

\subsection{Time complexity}\label{Sec:4.2}
LSTM is local in space and time \cite{hochreiter1997long}. Its computational complexity per time step and weight is $O(1)$, while the overall complexity of an LSTM per time step is equal to $O(w)$, where $w$ is the number of weights \cite{tsironi2017analysis}. Therefore, theoretically the computational complexity is proportion to the number of parameters in LSTM models. For activation functions, each of them will include 2 parameters, and will process one input for either forward or backward path. Therefore, the extra computational complexity will still be proportional to the number of activation parameters, which corresponds to the ratios shown in Table ~\ref{tab:3.1} and ~\ref{tab:3.1a}.  

For convolutional neural networks, we have learned that the total time complexity of all convolutional layers is \cite{he2015convolutional}:
\begin{equation}
    T(n) = O(\sum_{l=1}^d n_{l-1}\cdot s_l^2\cdot n_l\cdot m_l^2)
\end{equation}
where $l$ is the index of a convolutional layer, and $d$ is the
depth (number of convolutional layers). $n_l$ is the number of filters (also known as “width”) in the $l$-th layer, while $n_{l-1}$ is known as the number of input channels of the $l$-th layer. $s_l$ is the spatial size (length) of the filter. $m_l$ is the spatial size of the output feature map. For the whole training process we have:
\begin{equation}
O (m \cdot n_{iter} \cdot \sum^d_{l=1} (n_{l-1}\cdot s^2_l\cdot n_l \cdot m^2_l)) 
\end{equation}
where $m$ is the input length and $n_{iter}$ the number of iterations. When flexible activation function is implemented, there will be extra computational cost in both forward and backward propagations: In forward propagation, when flexible activation is added after a convolutional layer, the number of different activations in this layer equals to the number of channels in that convolutional layer. Therefore, we need to do more element-wise operations that is proportional to the number of the input elements to each flexible activation function. It depends on the number of input channels and input filter size to the activation layer, but does not depend on the number of output filters and corresponding filter size. This is because after the input mapping and activation function, each element will become a single value before multiplying with filters in the next layer. The extra computational cost can be written as:
\begin{equation}
    O (m \cdot n_{iter} \cdot \sum^d_{l=2} (n_{l-1}\cdot s^2_l)) 
\end{equation}
where we start from $l=2$ since the first activation layer comes after the first convolutional layer rather than the input layer of data. On the other hand, in backward propagation, the updating in Eq.~\eqref{Eq:2} will require extra computation that is proportional to the number of input element as well as the number of activation parameters, also we need to pass the gradients through flexible activation functions, which requires extra element-wise operations that proportional to the number of elements of the input to each flexible activation function. Notice that the gradients of the loss with respect to the output of activation are stored in cache and shared for calculating other gradients. 
Therefore, the theoretical extra computational cost is at least one-order smaller than that of the baseline model with fixed activation functions. This will not add a heavy burden for model training from the theoretical prospective.

\section{Conclusion} \label{Sec:5}
In this study, we proposed a set of principles for designing flexible activation functions in a weighted combination form. Based on these principles, we built two novel flexible activation functions: The first can be implemented to replace sigmoid and tanh functions in the RNN cells with bounded domains. The second is a combined form of ReLU and ELU family, which can be used as a substitute for activations with unbounded domain. In addition, two regularization terms considering the nature of layer-wise feature extraction and goodness of original activation functions are proposed, which is essential in controlling the model flexibility and achieving stable improvement of the models. 

Experiments on multivariate time series forecasting show that, with replacing sigmoid activation by the flexible combination proposed in this study, stacked LSTMs can achieve significant improvement in terms of convergence as well as validation performance for forecasting G7 Indices returns. On the other hand, the newly proposed P-E2-ReLU and P-E2-Id can stably outperform existing state-of-the-art activation functions including PReLU and ELU in image compression tasks with convolutional auto-encoder. In addition, we show that the towards-mean regularization on the flexible activation functions outperform the towards-default ones in general, and optimized regularization coefficients can further improve model performance on validation set.

Theoretically, by introducing combined flexible activation functions in a general form, it is not necessary to consider the types of single activation function in each layer as a hyper-parameter to be searched given the boundary of the output of each layer, and the shape of activation functions can be trained with back-propagation. In practice, the findings in this study provide a novel aspect to be considered in designing and improving deep learning architecture. The proposed flexible activation function in RNN cells could be a powerful tool in financial time series forecasting, which can also be used in combination with other advanced forecasting approaches.

In future study, different types of combined activation functions following the principles proposed in this study can be investigated on a variety of learning tasks. It can be implemented along with other methods in both meta-learning and manually designed architectures. Since we can have arbitrary number of components and activation parameters in the combined activation functions, this can be considered as another dimension of parameters in training deep neural networks. The corresponding approaches of learning the shapes of activations other than back-propagation can also be explored. 

\bibliography{IEEEabrv,research}
\bibliographystyle{IEEEtran}

\end{document}